# SAGE-Drive: Benefit-Aware Adaptive Reasoning for Autonomous Driving Planning

Donglei Rong, Wenbin Yao, Congcong Bai, Xi Gao, Qiaoqiao Ren, Sheng Jin, Cheng Xu

*Abstract*—**Explicit reasoning enables autonomous vehicles (AVs) to interpret complex traffic interactions and support trajectory planning, but introduces additional computational demands. Determining when explicit reasoning can improve planning accuracy remains a key challenge, as its benefits vary across driving scenarios. To address this challenge, this study proposes Scene-Adaptive Gating of Explicit Reasoning for Driving (SAGE-Drive), a benefit-aware adaptive reasoning framework for autonomous driving planning. First, paired planning outcomes with and without explicit reasoning are analyzed to measure the reasoning benefit, i.e., the reduction in planning error achieved by explicit reasoning, which serves as the supervision label. SAGE-Drive then learns to predict these benefits using geometric features of the initial trajectory planned without reasoning and multimodal scene information, and invokes explicit reasoning only for the scenarios with the highest predicted benefits under a predefined reasoning budget. Experiments on 1,497 paired driving scenarios reveal that explicit reasoning improves planning accuracy in 53.5% of cases while degrading it in 46.5%, yielding a near-zero average benefit. On the test set, with a 20% reasoning budget, SAGE-Drive achieves an average displacement error (ADE) of 1.5191 m, compared with 1.5252 m for random selection. At a 35% budget, it achieves an ADE of 1.5148 m, close to the 1.5141 m obtained when reasoning is applied to every scenario. These findings establish the predicted reasoning benefit as an effective criterion for reasoning selection, allowing explicit reasoning to be invoked where it improves planning accuracy at a fraction of its computational cost.**

*Index Terms*—**Autonomous driving, adaptive reasoning, benefit prediction, computational efficiency, trajectory planning.**

This study was supported by the National Key R&D Program of China (52672449) (*Corresponding author: Cheng Xu*)

Donglei Rong is with the School of Civil and Environmental Engineering, Georgia Institute of Technology, Atlanta GA 30324, USA (e-mail: drong8@gatech.edu)

Wenbin Yao is with the School of Civil Engineering and Architecture, Zhejiang Sci-Tech University, Hangzhou 310018, China (e-mail: wby@zstu.edu.cn)

Congcong Bai is with the Ocean College, Zhejiang University, Zhoushan 316021, China (e-mail: 12112125@zju.edu.cn).

Xi Gao is with the Institute of Intelligent Transportation Systems, College of Civil Engineering and Architecture, Zhejiang University, Hangzhou 310058, China and also with Department of Landscape Architecture and Environmental Planning, University of California, Berkeley, CA, 94720, USA (e-mail: 12312161@zju.edu.cn)

Qiaoqiao Ren is with the Transport Strategy Centre, Department of Civil and Environmental Engineering, Imperial College London, London SW7 2AZ, United Kingdom (e-mail: qiaoqiao.ren@imperial.ac.uk)

Sheng Jin is with the Institute of Intelligent Transportation Systems, College of Civil Engineering and Architecture, Zhejiang University, Hangzhou 310058, China (e-mail: jinsheng@zju.edu.cn)

Cheng Xu is with the Department of Traffic Management Engineering, Zhejiang Police College, Hangzhou 310053, China (xucheng@zjjcxy.cn)

## I. Introduction

AUTONOMOUS vehicles (AVs) operate in diverse traffic environments. Rule-based and learning-based planning methods perform reliably in routine situations, where road layouts, traffic rules, and interactions with other road users are predictable. In complex scenarios, however, such as those involving unexpected behavior of other agents, limited visibility, or ambiguous right-of-way, these methods may not capture how traffic participants interact, leading to unsafe maneuvers or unnecessary delays. Handling such scenarios requires an AV to understand the scene and its interactions before an action. Vision-language models (VLMs) offer a natural way to meet this need: they can generate intermediate reasoning steps that interpret visual observations, infer interactions among road users, and evaluate candidate driving actions. This explicit reasoning enables an AV to identify potential conflicts, anticipate the consequences of alternative maneuvers, and select trajectories that lower collision risk while avoiding unnecessary disruptions to traffic flow.

Existing applications of explicit reasoning in autonomous driving (AD) fall into three categories: scene understanding, reasoning-guided planning, and adaptive reasoning. For scene understanding, DriveLM [1] uses graph-structured question answering to link perception, prediction, and planning. Reason2Drive [2] introduces annotated reasoning chains to support multi-step interpretation of driving situations. For reasoning-guided planning, DriveVLM [3] integrates scene description, scene analysis, and hierarchical planning. VLR-Driver [4] uses spatiotemporal reasoning to analyze traffic interactions and generate driving decisions. More recently, adaptive reasoning methods adjust whether and how much to reason according to driving conditions. FactorDrive [5] adapts reasoning steps based on planning-critical factors, and STEER [6] learns to choose between direct prediction and explicit reasoning using planning-related rewards. These studies show that explicit reasoning supports scene understanding and driving decisions, and that its computational cost motivates more efficient reasoning strategies. However, none of these methods explicitly estimates how much reasoning will improve or degrade trajectory planning accuracy before invoking it. As a result, reasoning can be invoked in routine scenes, where it adds latency or even degrades the trajectory, and skipped in challenging interactions, where it would improve the driving decision.

To address this gap, this study proposes Scene-Adaptive Gating of Explicit Reasoning for Driving (SAGE-Drive), a benefit-aware adaptive reasoning framework for AD planning. First, the reasoning benefit is defined as the reduction in trajectory planning error achieved by explicit reasoning. It is measured from paired planning results with and without reasoning and serves as the supervision for benefit prediction. SAGE-Drive then combines multimodal scene representations from camera observations, LiDAR data, and ego-vehicle motion history with geometric features of an initial trajectory generated without explicit reasoning. Based on these features, a benefit predictor estimates the reasoning benefit before explicit reasoning is invoked. Under a predefined reasoning budget, SAGE-Drive selects the scenarios with the highest predicted benefits for explicit reasoning and retains the initial trajectories for the rest.

The remainder of this paper is organized as follows. Section II reviews related work on scene understanding, reasoning-guided planning, and adaptive reasoning for AD. Section III presents the proposed SAGE-Drive framework, including reasoning benefit supervision, multimodal scene representation, and budget-constrained reasoning selection. Section IV describes the experimental setup and evaluates planning accuracy, computational efficiency, and the effects of different reasoning budgets. Finally, Section V concludes the paper.

## II. Related Work

Related work on explicit reasoning for AD falls into three categories: scene understanding, reasoning-guided planning, and adaptive reasoning. These categories build on one another, progressing from interpreting traffic scenes and their interactions to using such understanding for driving decisions and trajectory planning, and finally to deciding whether and how much to reason during planning (Fig. 1).

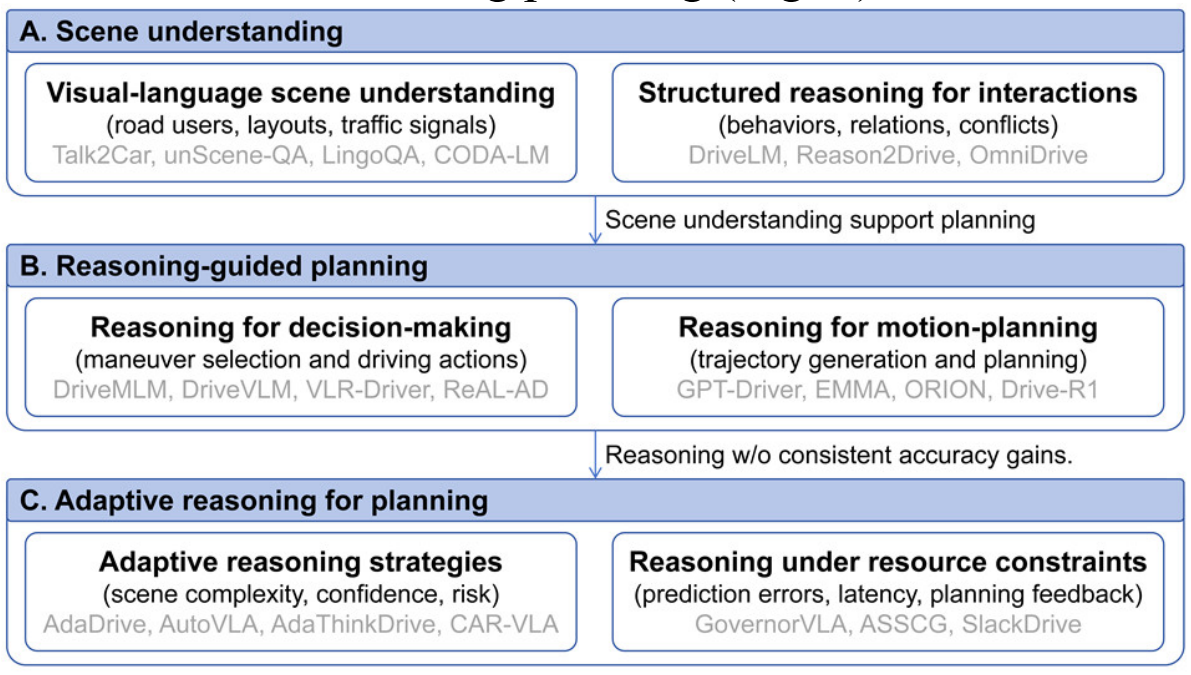

**Fig. 1.** Framework of related work.

### *A. Scene Understanding*

VLMs have been applied to AD to interpret visual observations and describe road users, road layouts, and traffic signals in natural language. Talk2Car [7] investigates visual grounding by identifying objects in traffic scenes from natural-language driving commands. nuScenes-QA [8] combines camera images and LiDAR data to support question answering about surrounding objects and their spatial relationships. LingoQA [9] extends visual question answering to driving videos, enabling the interpretation of traffic scenes across consecutive frames. CODA-LM [10] further examines the ability of VLMs to describe traffic scenes and identify unusual objects in corner-case driving scenarios. These studies establish the use of visual and language information to represent the key elements of traffic environments.

Beyond identifying scene elements, recent studies have explored structured reasoning to understand interactions among road users. DriveLM [1] uses graph-structured visual question answering to link object identification, traffic interactions, and behavior prediction. Reason2Drive [2] organizes driving scene interpretation into sequential perception, prediction, and reasoning steps to examine road-user behaviors and their implications for driving decisions. NuScenes-SpatialQA [11] uses 3D scene graphs to support reasoning about relative positions and distances among road users. OmniDrive [12] further introduces counterfactual reasoning to assess how alternative driving actions would affect future traffic situations.

These advances in scene understanding provide essential information about road users, traffic interactions, and potential conflicts. Reasoning-guided planning uses this information to evaluate candidate maneuvers and generate trajectories that account for surrounding traffic conditions.

### *B. Reasoning-Guided Planning*

Reasoning-guided planning methods incorporate explicit reasoning directly into driving decision-making. DriveMLM [13] aligns multimodal language model outputs with behavioral planning states, enabling driving decisions based on sensor observations, traffic rules, and user instructions. DriveVLM [3] integrates scene description, scene analysis, and hierarchical planning to link traffic scene interpretation with driving decisions. VLR-Driver [4] uses spatiotemporal chain-of-thought reasoning to assess potential risks and infer the intentions of surrounding road users before driving actions. ReAL-AD [14] introduces hierarchical reasoning that organizes decision-making into strategic, tactical, and operational levels. These studies show that explicit reasoning translates traffic scene understanding into driving decisions that guide motion planning.

Explicit reasoning has also been integrated into motion planning. GPT-Driver [15] reformulates motion planning as a language modeling task and uses a fine-tuning strategy to generate trajectory. EMMA [16] uses a multimodal language model to process camera observations and navigation commands, generating future trajectories through a unified language representation. ORION [17] links scene reasoning with trajectory generation by introducing a planning token that conditions a generative planner on the interpreted driving context. More recently, Drive-R1 [18] incorporates reinforcement learning to align reasoning processes with motion planning outcomes, using rewards derived from predicted trajectories and driving actions. These studies integrate explicit reasoning into trajectory generation, linking traffic scene interpretation with the prediction of future vehicle motion.

However, applying explicit reasoning uniformly across driving scenarios adds computational cost without consistently improving planning accuracy, motivating adaptive reasoning strategies that determine when reasoning is beneficial.

### C. Adaptive Reasoning for Planning

Adaptive reasoning strategies adjust whether and how much to reason according to each driving scenario. AdaDrive [19] introduces an adaptive slow-fast framework that determines when to activate language-model reasoning based on scene complexity and prediction confidence. AutoVLA [20] combines fast trajectory prediction with slow chain-of-thought reasoning, using reinforcement fine-tuning to reduce unnecessary reasoning in routine scenarios. AdaThinkDrive [21] develops an adaptive thinking mechanism that learns to select between direct prediction and explicit reasoning with a reward that compares trajectory quality across the two modes. ThinkDrive [22] uses scene complexity estimation to trigger asynchronous causal reasoning while maintaining real-time planning. More recently, CAR-VLA [23] jointly considers scene complexity and dynamic risk to adjust reasoning depth, urgency, and focus, distinguishing between direct prediction, deliberate reasoning, and hazard-focused responses. These approaches decide when to reason by scene complexity, prediction confidence, and risk, or learn this decision from trajectory-based rewards.

Another line of work allocates reasoning computation to balance planning accuracy and inference latency. GovernorVLA [24] introduces a learned routing mechanism that identifies driving situations with high predicted trajectory errors and selectively activates a more computationally expensive reasoning policy. ASSCG [25] formulates reasoning invocation as a resource-aware sequential decision problem, making frame-level query, cache, and drop decisions to refresh, reuse, or suppress slow-system guidance. SlackDrive [26] addresses runtime computational constraints by selecting the visual-token input budget before each inference based on available processing time and estimated planning performance. These approaches advance reasoning allocation by incorporating prediction errors, computational constraints, and planning feedback into inference decisions.

Existing adaptive reasoning methods allocate computation based on scene complexity, model uncertainty, computational constraints, or planning-related feedback. However, these factors provide no estimate, before reasoning is invoked, of how much explicit reasoning will improve or degrade trajectory planning accuracy relative to direct planning. This motivates benefit-aware reasoning allocation that estimates the expected reasoning benefit in advance and uses this estimate to guide reasoning selection under a predefined reasoning budget.

## III. Methodology

This section presents SAGE-Drive, a benefit-aware framework that selectively invokes explicit reasoning to improve trajectory planning accuracy. The framework first measures the reasoning benefit, defined as the reduction in trajectory planning error achieved by explicit reasoning, from paired planning trajectories generated with and without reasoning. These measurements serve as supervision for benefit prediction. SAGE-Drive then extracts multimodal scene representations from camera observations, LiDAR data, ego-vehicle motion history, and navigation information, together with geometric features derived from an initial trajectory generated without explicit reasoning. Based on these features, a benefit predictor estimates the reasoning benefit before explicit reasoning is invoked. Under a predefined reasoning budget, SAGE-Drive selects the scenarios with the highest predicted benefits for explicit reasoning and retains the initial trajectories for the rest (Fig. 2).

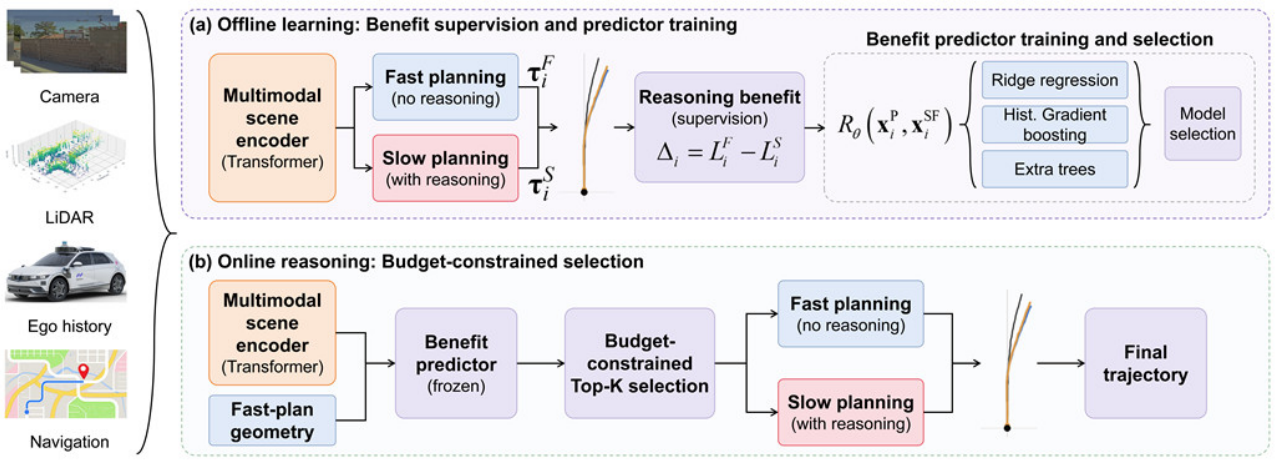

**Fig. 2.** SAGE-Drive framework.

### A. Problem Formulation

(1) **Reasoning benefit**. A driving scenario is described by four inputs: camera observations, LiDAR data, ego-vehicle motion history, and navigation information. Given these inputs, fast planning directly generates a future ego-vehicle trajectory, whereas slow planning incorporates explicit reasoning into trajectory generation. Planning error is measured by the average displacement error (ADE), the mean Euclidean distance between planned and ground-truth trajectories over the entire planning horizon. The reasoning benefit is defined as the ADE of fast planning minus that of slow planning, so a positive value indicates that explicit reasoning reduces planning error. The definitions are shown below:

$$L_i^m = \frac{1}{T}\sum_{t=1}^{T}\left\| \mathbf{p}_{i,t}^m - \mathbf{p}_{i,t}^* \right\|_2, m \in \{F, S\}, \tag{1}$$

$$\Delta_i = L_i^F - L_i^S, \tag{2}$$

where $i$ denotes the driving scenario. $m \in \{F, S\}$ denotes fast and slow planning, respectively. $\mathbf{p}_{i,t}^m$ denotes the predicted ego-vehicle position at future time step $t$. $\mathbf{p}_{i,t}^*$ denotes the ground-truth ego-vehicle position at future time step $t$. $T$=50 denotes the number of future trajectory points sampled at 10 Hz. $\Delta_i$ denotes the reasoning benefit for scenario $i$. $L_i^F$ and $L_i^S$ denote the planning errors under fast and slow planning, respectively.

(2) **Reasoning allocation**. Reasoning allocation across driving scenarios is formulated with a binary decision variable that indicates whether slow planning is invoked for each scenario. Given a predefined reasoning budget, the objective is to minimize the mean ADE across scenarios by selecting scenarios that provide the greatest reasoning benefit. This selection problem is equivalent to maximizing the total reasoning benefit under a fixed number of slow planning calls:

$$\begin{aligned} &\max_{a} \sum_{i=1}^{N} a_i \Delta_i \\ &\text{s.t.} \sum_{i=1}^{N} a_i = K, \\ &\quad a_i \in \{0,1\}, i = 1, \dots, N, \\ &\quad K = \text{round}(bN), \end{aligned} \tag{3}$$

where $\mathbf{a} = [a_1, \ldots, a_N]$ denotes the reasoning selection vector, with $a_i = 1$ indicating slow planning and $a_i = 0$ indicating fast planning. $N$ denotes the number of driving scenarios. $b \in (0, 1]$ denotes the predefined proportion of slow planning calls. $K$ denotes the corresponding number of scenarios selected for explicit reasoning.

### *B. Reasoning Benefit Supervision*

(1) **Paired planning outcomes**. Supervision for benefit prediction is constructed from paired planning outcomes, obtained by applying fast and slow planning to the same driving scenario using an identical planning backbone. The resulting trajectories, denoted $\boldsymbol{\tau}_i^F = \left\{\mathbf{p}_{i,1}^F, \ldots, \mathbf{p}_{i,T}^F\right\}$ and $\boldsymbol{\tau}_i^S = \left\{\mathbf{p}_{i,1}^S, \ldots, \mathbf{p}_{i,T}^S\right\}$, are matched using unique scenario identifiers to ensure that both trajectories correspond to the same traffic observations and ground-truth trajectory. Each trajectory covers a 5 s horizon with 50 positions sampled at 10 Hz, and the paired outcomes are validated for trajectory completeness, finite coordinates, and consistency of ground-truth positions.

(2) **Benefit supervision dataset**. The paired planning outcomes form a supervised dataset for learning the reasoning benefit. For each driving scenario, the planning errors of fast and slow planning are calculated using Eq. (1), and their difference, defined in Eq. (2), serves as the continuous benefit label. The resulting dataset contains both positive and negative benefit values, allowing the benefit predictor to learn whether and by how much explicit reasoning reduces or increases trajectory planning error. The supervision dataset is formulated as:

$$\mathbf{D}_{\text{sup}} = \left\{\mathbf{o}_i, \boldsymbol{\tau}_i^F, \Delta_i\right\}_{i=1}^N, \tag{4}$$

where $\mathbf{D}_{\text{sup}}$ denotes the reasoning benefit supervision dataset. $\mathbf{o}_i$ denotes the multimodal observations of driving scenario $i$.

### *C. Multimodal Scene Representation*

(1) **Multimodal feature extraction**. Multimodal features relevant to benefit prediction are extracted from camera observations, LiDAR point clouds, ego-vehicle motion history, and navigation information. Images from eight surround-view cameras at four time steps are processed by a frozen DINOv2-Small encoder, producing 32 visual feature vectors of 384 dimensions each. LiDAR point clouds are represented by 24-dimensional spatial features, including point distributions across radial and angular regions, point density, elevation statistics, and average distance. Ego-vehicle motion history is encoded as an 80-dimensional vector derived from 16 historical and current states, each describing two-dimensional relative position, heading, speed, and yaw rate. Navigation commands are represented by four-dimensional one-hot vectors indicating left turns, right turns, straight or following movements, and other commands.

(2) **Multimodal feature fusion**. A Transformer-based fusion encoder integrates the extracted multimodal features into a unified representation of the driving scene. Visual features are projected into a 192-dimensional latent space and supplemented with camera-view and temporal embeddings to preserve their spatial and temporal information. LiDAR, ego-motion, and navigation features are projected into the same latent space and assigned modality embeddings. A learnable [CLS] token is concatenated with these features and processed by a two-layer Transformer encoder with four attention heads to capture dependencies across modalities. The output [CLS] token is layer-normalized to obtain the scene representation:

$$\mathbf{z}_i = E_\phi\left(\mathbf{o}_i\right), \tag{5}$$

where $E_\phi$ denotes the multimodal fusion encoder parameterized by $\phi$. $\mathbf{z}_i$ denotes the resulting L2-normalized scene embedding.

(3) **Self-supervised representation learning**. The fusion encoder is trained through self-supervised masked visual feature reconstruction to learn relationships among different sensor modalities and temporal observations. During training, 30% of the available visual tokens are randomly masked, and the current-frame visual features are additionally masked for 40% of the samples. A linear reconstruction head predicts the masked visual features from the remaining multimodal information, using cosine distance between reconstructed and original DINOv2 features as the reconstruction loss. To prevent representation collapse, a variance regularization term is applied to the scene representations. The overall training objective is formulated as:

$$L_{\text{SSL}} = \underbrace{\frac{1}{M} \sum_{(i,j) \in M} \left(1 - \frac{\bar{\mathbf{v}}_{i,j}^{\mathrm{T}} \mathbf{v}_{i,j}}{\left\|\bar{\mathbf{v}}_{i,j}\right\|_2 \left\|\mathbf{v}_{i,j}\right\|_2}\right)}_{L_{\text{rec}}} + \lambda_{\text{var}} L_{\text{var}}, \tag{6}$$

where $M$ denotes the set of masked visual tokens in a training batch. $\mathbf{v}_{i,j}$ and $\bar{\mathbf{v}}_{i,j}$ denote the original and reconstructed visual feature vectors of token $j$ in scenario $i$, respectively. $L_{\text{rec}}$ denotes the cosine reconstruction loss. $L_{\text{var}}$ penalizes embedding dimensions whose batch-wise standard deviation falls below 0.06 [27]. $\lambda_{\text{var}}$ controls the regularization strength.

### *D. Plan-Conditioned Benefit Prediction*

(1) **Fast-plan geometry encoding**. To characterize the initial planning behavior, geometric and kinematic features are extracted from the fast planning trajectory. From its 50 planned positions and the initial ego-vehicle speed, motion characteristics including velocity, acceleration, jerk, heading changes, and lateral motion are computed. These measurements are organized into a 67-dimensional feature vector, comprising 31 global features that describe the overall motion profile, endpoint displacement, and path geometry, and 36 horizon-specific features, six at each planning horizon of 0.5, 1, 2, 3, 4, and 5 s. The resulting fast-plan geometry representation is defined as:

$$\mathbf{g}_i = \Phi\left(\boldsymbol{\tau}_i^F, v_{i,0}\right), \tag{7}$$

where $\Phi(\cdot)$ denotes the geometric feature extraction function. $\boldsymbol{\tau}_i^F$ denotes the initial trajectory generated by fast planning for scenario $i$. $v_{i,0}$ denotes the initial ego-vehicle speed. $\mathbf{g}_i$ denotes the results fast-plan geometry feature vector.

(2) **Reasoning benefit prediction**. To estimate the reasoning benefit before slow planning is invoked, two candidate feature representations are constructed from the fast-plan geometry features and the multimodal scene embedding. The plan-only representation uses the geometry features $\mathbf{g}_i$, while the scene-fusion representation combines these features with the scene embedding $\mathbf{z}_i$. The scene embedding is standardized and reduced to 12 dimensions using principal component analysis (PCA), then concatenated with the standardized fast-plan geometry features. The reasoning benefit prediction score are expressed as follows:

$$\mathbf{x}_i^{\mathrm{P}} = \mathbf{g}_i^{'}, \mathbf{x}_i^{\mathrm{SF}} = \left[\mathbf{g}_i^{'}, P_{12}\left(\mathbf{z}_i^{'}\right)\right], \tag{8}$$

$$s_i = R_\theta\left(\mathbf{x}_i^{\mathrm{P}}, \mathbf{x}_i^{\mathrm{SF}}\right), \tag{9}$$

where $\mathbf{x}_i^{\mathrm{P}}$ and $\mathbf{x}_i^{\mathrm{SF}}$ denote the plan-only and scene-fusion feature representations, respectively. $\mathbf{g}_i^{'}$ and $\mathbf{z}_i^{'}$ represent the standardized fast-plan geometry and scene embeddings. $P_{12}$ denotes PCA projection onto 12 principal components. $R_\theta$ represents the benefit scoring model. $s_i$ denotes the predicted benefit score.

(3) **Model training and selection**. The benefit predictors are trained using the reasoning benefit $\Delta_i$ as the supervision target. Five predictor variants are considered: Ridge Regression and Histogram Gradient Boosting using both plan-only and scene-fusion features, and Extra Trees Regression using plan-only features. An additional ensemble candidate combines the percentile ranks of predictions from scene-fusion Ridge Regression, plan-only Histogram Gradient Boosting, and plan-only Extra Trees Regression. All feature preprocessing operations, including standardization and PCA, are fitted exclusively on the training data.

The benefit predictor is selected by three-fold cross-validation grouped by driving log, which yields out-of-fold predictions for all candidate models. Model performance is evaluated by the reduction in mean ADE relative to random reasoning selection under the same predefined reasoning budget. For candidate model (c), the improvement is calculated as:

$$U_c\left(b\right) = \frac{1}{N}\sum_{i=1}^{N} a_i^{(c)}\Delta_i - \frac{K}{N}\left(\frac{1}{N}\sum_{i=1}^{N}\Delta_i\right), \tag{10}$$

where $U_c(b)$ denotes the improvement over expected random selection at reasoning budget $b$. $a_i^{(c)}$ denotes whether scenario $i$ is among the $K$ highest-scoring scenarios ranked by candidate model $c$. The three candidates with the highest out-of-fold gains $U_c(b)$ are shortlisted, and the candidate achieving the greatest gain on the log-disjoint validation set is selected.

### E. Budget-Constrained Reasoning Selection

(1) **Reasoning allocation**. Based on the benefit scores predicted in Section III-D, SAGE-Drive selects driving scenarios for explicit reasoning according to their predicted reasoning benefit. For a predefined reasoning budget $b$, all scenarios are ranked in descending order of their predicted benefit scores, and the top $K$ scenarios are selected for Slow Planning, where $K$ follows from the budget constraint in Eq. (3).

(2) **Adaptive reasoning outputs**. After reasoning selection, Slow Planning is invoked for the selected scenarios, and the remaining scenarios retain their Fast Planning trajectories. The final trajectory for each scenario is therefore determined by the reasoning selection decision:

$$\boldsymbol{\tau}_i^{'} = z_i \boldsymbol{\tau}_i^{S'} + \left(1 - z_i\right)\boldsymbol{\tau}_i^{F'}, \tag{11}$$

where $\boldsymbol{\tau}_i^{'}$ denotes the final trajectory for scenario $i$. $\boldsymbol{\tau}_i^{S}$ and $\boldsymbol{\tau}_i^{F}$ denote the trajectories generated by fast and slow planning, respectively. $z_i$ denotes the final decision for planning, ($z_i = 1$ means slow planning, $z_i = 0$ means fast planning).

## IV. Experimental Analysis

### A. Experimental Setup

(1) Dataset description. The experiments are conducted on a multimodal paired-planning dataset constructed from nuReasoning [28], spanning 340 driving logs and 499 clips, with three frames sampled from each clip. Each driving scenario is represented by multi-view camera observations, LiDAR point clouds, ego-vehicle motion histories, and navigation commands. For each scenario, Fast and Slow Planning generate separate trajectories over a 5 s planning horizon, yielding 1,497 matched pairs. The pairs are split by driving log into training, validation, and test sets of 1,047, 225, and 225 pairs, respectively, so that samples from the same log remain in a single subset. Additional unlabeled frames from training clips are used for self-supervised pretraining of the fusion encoder; the validation and test sets are used for model selection and performance evaluation, respectively.

(2) **Baseline models**. To evaluate the effectiveness of benefit-aware reasoning selection, SAGE-Drive is compared with three baseline planning strategies: Fast-only, Slow-only, and Random Selection. All methods use Qwen-Drive-1.0-4B as the planning backbone so that performance differences arise only from reasoning selection. Fast-only applies Fast Planning to every scenario, whereas Slow-only applies Slow Planning to every scenario. Random Selection invokes Slow Planning for randomly selected scenarios under a predefined reasoning budget and retains Fast Planning trajectories for the remaining scenarios. SAGE-Drive instead selects scenarios according to their predicted reasoning benefits. For a consistent comparison, Random Selection and SAGE-Drive are evaluated under the same reasoning budget, with 20% as the primary setting. Planning performance is evaluated using the same paired trajectories and ground-truth references across all strategies.

(3) **Evaluation metrics**. Trajectory planning performance is evaluated using average displacement error (ADE) and final displacement error (FDE). ADE measures the average Euclidean distance between planned and ground-truth positions over the planning horizon, while FDE measures their displacement at the final planning step. Both metrics are reported in meters, with ADE50 and FDE50 serving as the primary evaluation metrics over a 5 s horizon. To examine temporal differences in planning accuracy, ADE is additionally evaluated over horizons of 10, 20, 30, and 40 planning steps, corresponding to 1 to 4 s. Reasoning efficiency is measured by the Slow Planning call rate, defined as the proportion of scenarios assigned to Slow Planning. The effectiveness of reasoning selection is assessed through ADE reductions relative to Fast-only and Random Selection. Statistical uncertainty is quantified using 95% confidence intervals obtained through driving-log-level cluster bootstrap resampling.

(4) **Implementation details**. The implementation configurations of SAGE-Drive, including model architecture,

training parameters, and reasoning budget settings, are summarized in Table I.

TABLE I
IMPLEMENTATION DETAILS OF SAGE-DRIVE

| Component | Configuration | Value |
|---|---|---|
| Planning model | Backbone | Qwen-Drive-1.0-4B |
| Visual encoder | Backbone | DINOv2-Small |
| Scene encoder | Architecture | Transformer |
| | Layers | 2 |
| | Attention heads | 4 |
| | Embedding dimension | 192 |
| Representation learning | Optimizer | AdamW |
| | Learning rate | 2×10–4 |
| | Weight decay | 0.03 |
| | Batch size | 32 |
| | Training epochs | 12 |
| | Visual making ratio | 30% |
| Benefit prediction | Fast-plan geometry | 67 dimensions |
| | Scene embedding after PCA | 12 dimensions |
| | Scene-fusion features | 79 dimensions |
| | Cross-validation | 3-fold, log-grouped |
| Reasoning allocation | Default slow budget | 20% |
| | Budget sweep | 0%-100% |
| | Budget interval | 5% |
| | Selection strategy | Top-K benefit ranking |

### *B. Reasoning Benefit Analysis*

To investigate the effects of explicit reasoning on trajectory planning accuracy, paired fast and slow planning outcomes are analyzed from three perspectives: overall reasoning benefit, temporal benefit variation, and scene-dependent reasoning benefit.

(1) **Overall reasoning benefit**. As shown in Fig. 3, explicit reasoning has heterogeneous effects across the 1,497 paired driving scenarios. It reduces ADE in 53.5% of scenarios and increases it in the remaining 46.5%, with large variation in the size of these changes. The direction of the change is largely consistent along the trajectory: ADE and FDE benefits share the same sign in 84.3% of scenarios, so gains and losses generally persist to the final position. These findings show that the effect of explicit reasoning is heterogeneous and motivate selective reasoning that accounts for both its gains and its losses.

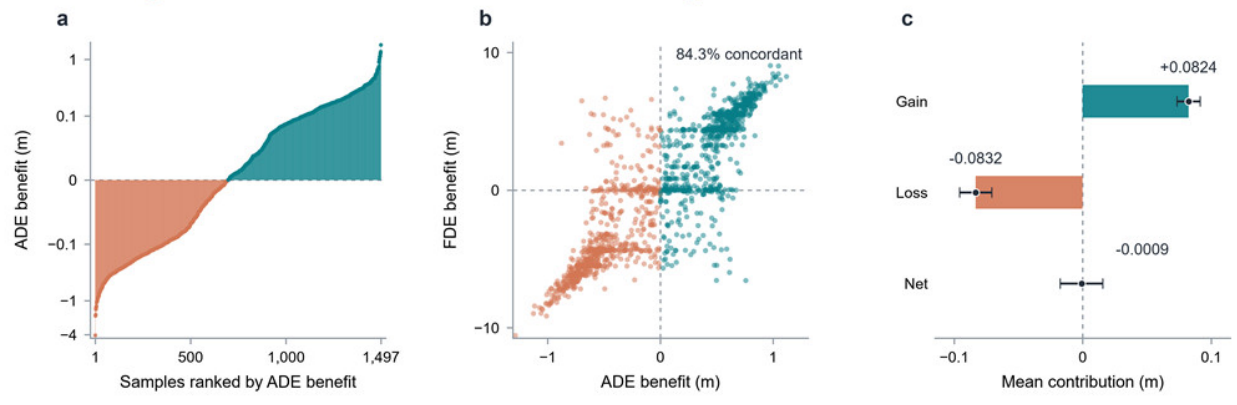

**Fig. 3.** Overall reasoning benefit.

(2) **Temporal benefit variation**. The benefit of explicit reasoning is concentrated at the beginning of the planning horizon (Fig. 4). Fast and Slow Planning accumulate position error at almost the same rate over the 50 steps (Fig. 4(a)), so their pointwise difference stays small: it is positive at early steps but fluctuates around zero with widening uncertainty as the horizon extends (Fig. 4(b)). The cumulative benefit accordingly rises at first and then decays toward zero (Fig. 4(c)). Aggregated by interval, the benefit is clearest within the first 10 steps, whereas later intervals have wider uncertainty ranges that overlap zero (Fig. 4(d)). Explicit reasoning therefore yields a small short-term gain but no sustained improvement in long-horizon trajectory accuracy.

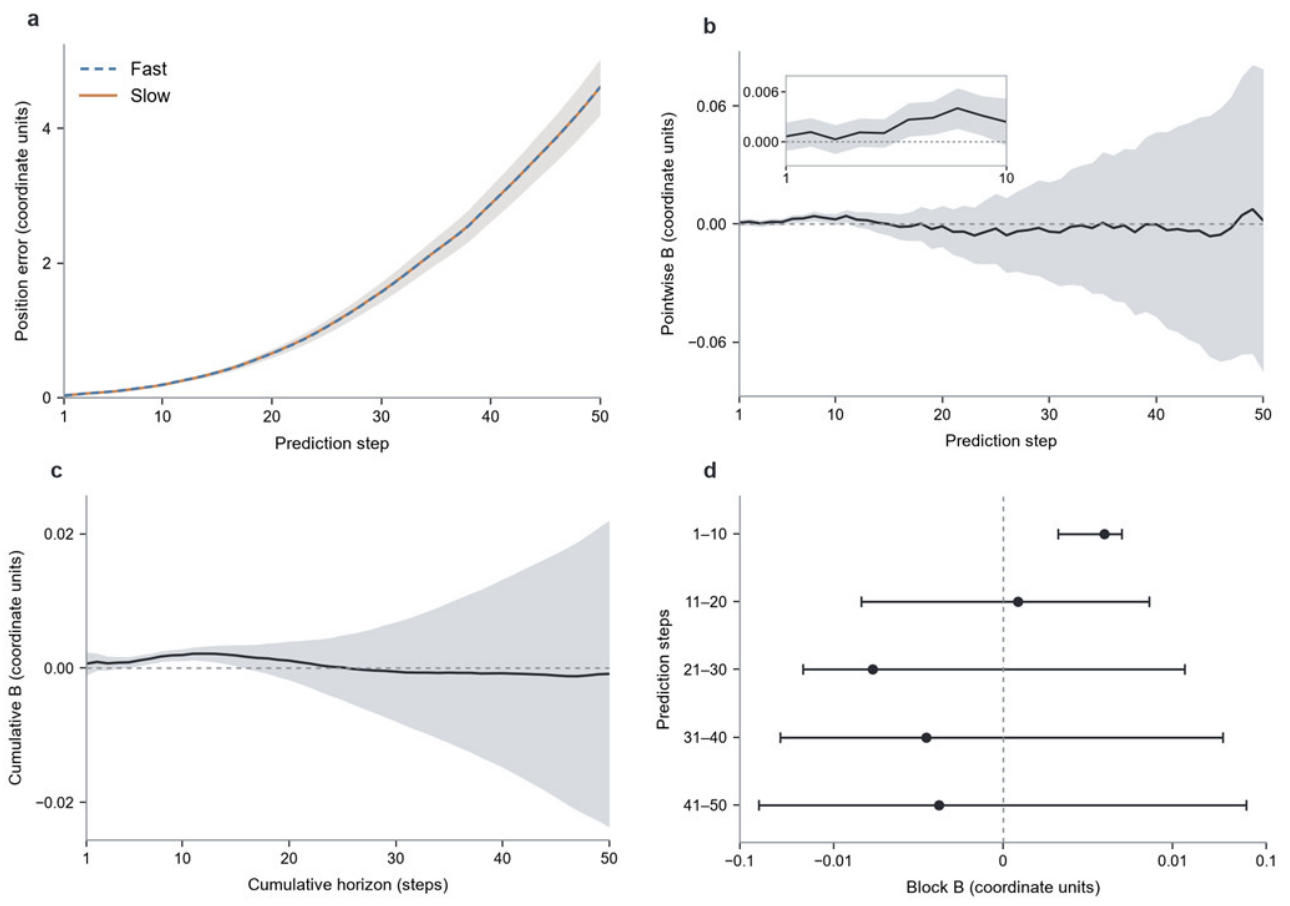

**Fig. 4.** Temporal benefit variation.

(3) **Scene-dependent reasoning benefit**. As shown in Fig. 5, neither initial speed nor scene identity reliably indicates whether explicit reasoning will improve planning accuracy. The mean benefit varies nonlinearly with initial speed, being slightly negative at intermediate speeds and positive at low and high speeds, but its uncertainty intervals largely overlap zero. Scene-adjusted estimates show the same pattern. The benefit also varies within individual scenes: across three sampled frames, most scenes contain both gains and losses, and a concordance test with a p-value of 0.957 shows that the within-scene consistency of the benefit is no higher than chance. These results motivate estimating the reasoning benefit from frame-level features rather than from coarse scene attributes.

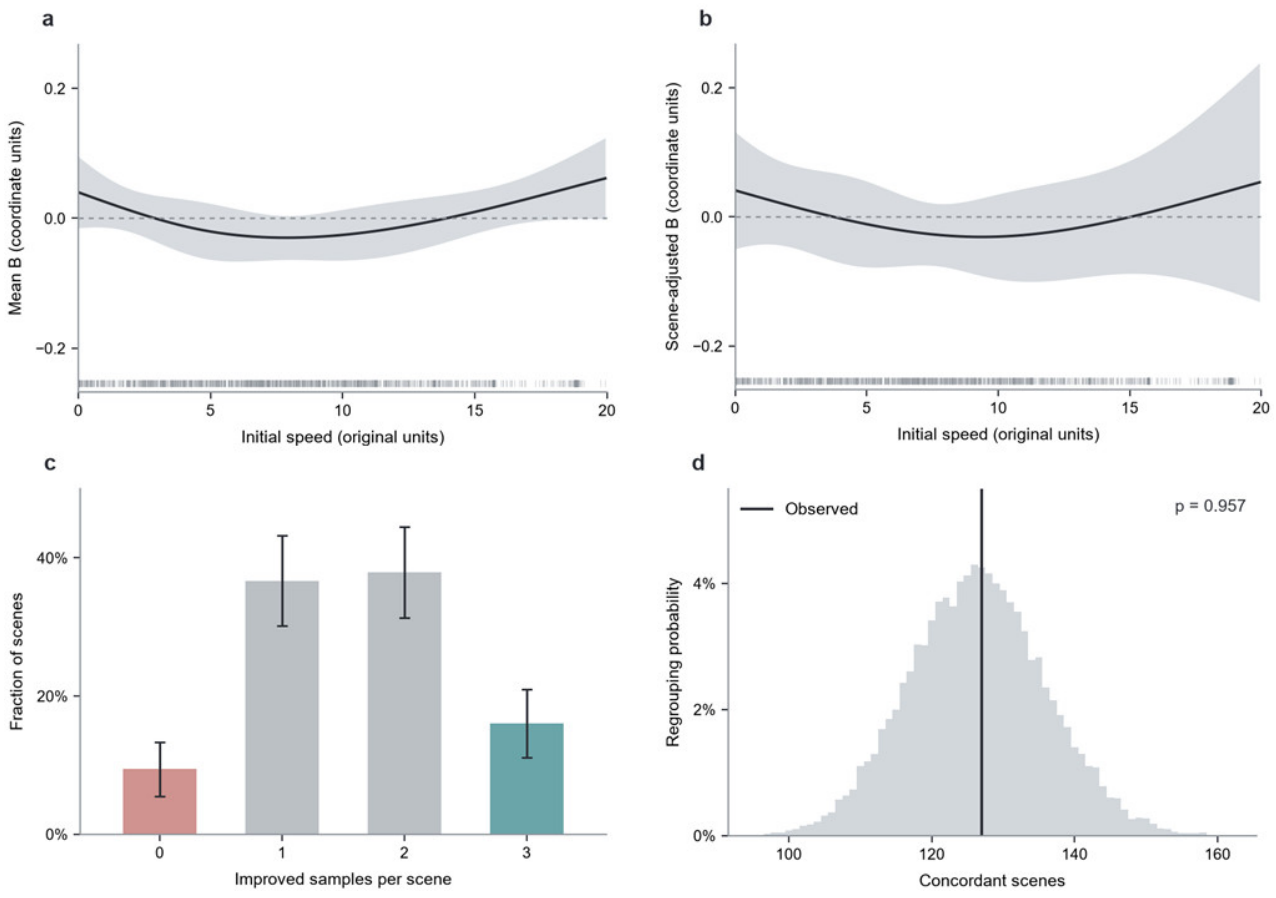

**Fig. 5.** Scene-dependent reasoning benefit.

### *C. Planning Performance Analysis*

Table II compares the trajectory planning performance of SAGE-Drive with three baselines. Slow-only achieves the lowest ADE and FDE of 1.5141 m and 4.6032 m, respectively, compared with 1.5279 m and 4.6590 m for fast-only. With a 20% reasoning budget, SAGE-Drive reduces ADE from 1.5252 m to 1.5191 m and FDE from 4.6479 m to 4.6315 m relative to random selection under the same budget. SAGE-Drive achieves approximately 64% of the ADE improvement obtained by slow-only while invoking slow planning for only 20% of scenarios. Therefore, benefit-guided reasoning selection improves planning accuracy over random selection while using 80% fewer slow planning calls than slow-only.

TABLE II
PLANNING PERFORMANCE COMPARISON

| Method | Slow calls (%) | ADE (m) ↓ | FDE (m) ↓ |
|---|---|---|---|
| Fast-only | 0.0 | 1.5279 | 4.6590 |
| Slow-only | 100.0 | 1.5141 | 4.6032 |
| Random | 20.0 | 1.5252 | 4.6479 |
| SAGE-Drive | 20.0 | 1.5191 | 4.6315 |

As shown in Fig. 6, SAGE-Drive narrows the gap between random selection and slow-only under the 20% reasoning budget, with a larger relative error reduction over random selection in ADE than in FDE. The realized benefits of the selected scenarios, ordered by predicted benefit, include both gains and losses, but several large gains appear among the highest-ranked scenarios. The improvement of SAGE-Drive therefore comes from ranking high-benefit scenarios near the top.

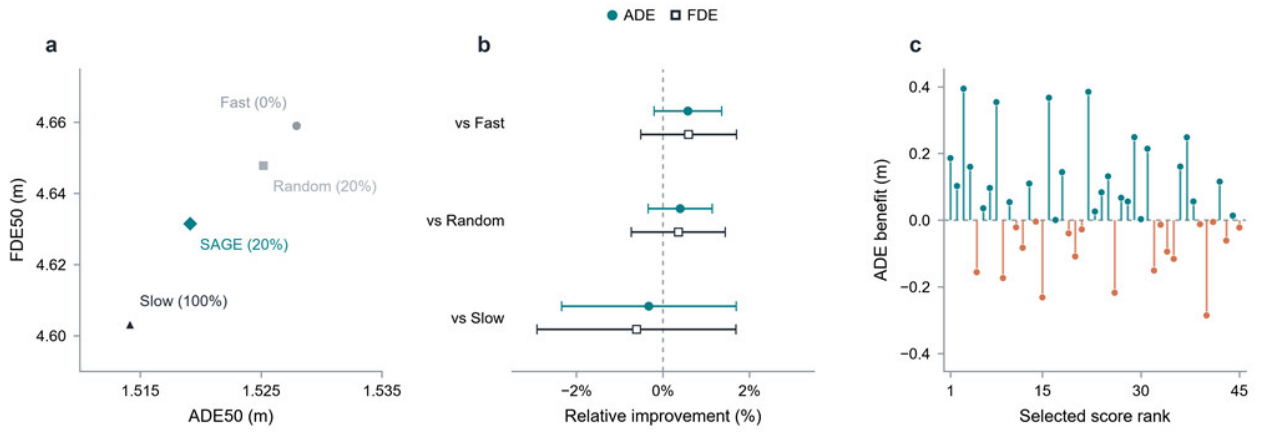

**Fig. 6.** Planning performance comparison.

## D. Reasoning Budget Analysis

Planning error decreases as the reasoning budget increases, and SAGE-Drive separates from random selection mainly at longer planning horizons, as shown in Fig. 7(a)–(e). At short horizons, the two curves nearly coincide. The SAGE-Drive curve is not monotonic at intermediate budgets because the realized benefits of the additionally selected scenarios vary. At a 35% reasoning budget, SAGE-Drive reaches an ADE of 1.5148 m over the full 50-step horizon, close to the 1.5141 m of Slow-only. This recovers about 95% of the ADE improvement of Slow-only over Fast-only with 35% of the Slow Planning calls.

The relative gain of SAGE-Drive over random selection depends on both the reasoning budget and the planning horizons. At short horizons, the gain stays close to zero at most budgets. At longer horizons, positive gains emerge at several low and intermediate budgets, but the uncertainty bands also widen because the benefits of the selected scenarios vary more over longer horizons.

Under the 20% reasoning budget, benefit-guided selection shifts the selected scenarios toward positive reasoning benefits, mainly at longer horizons. At short horizons, the distributions of selected and unselected scenarios nearly coincide. At longer horizons, the selected scenarios shift modestly toward positive benefits. Both groups still contain gains and losses, and the magnitude of both grows with the planning horizon.

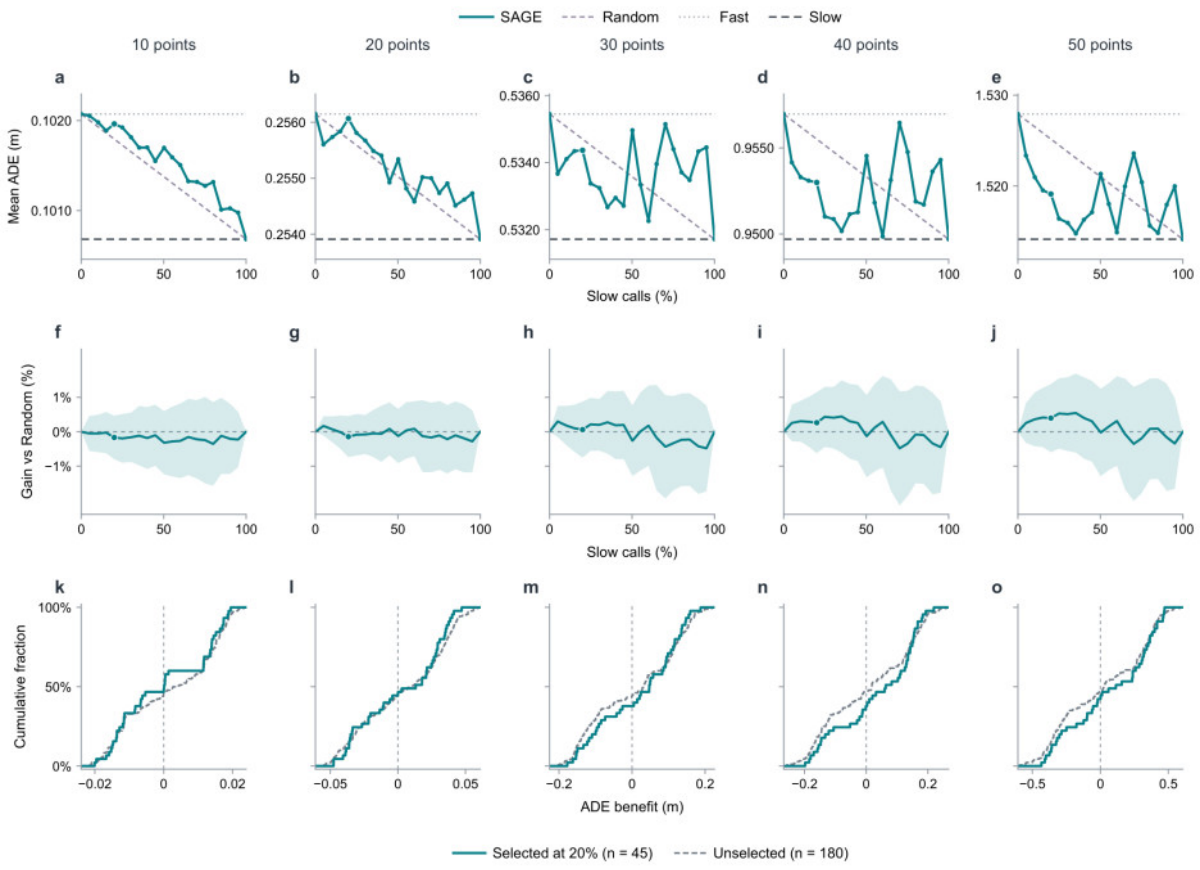

**Fig. 7.** Budget horizon atlas.

## E. Ablation Analysis

As shown in Fig. 8, the gain from reasoning selection depends on both the selection strategy and the computational constraints. ADE improvements vary across strategies and reasoning budgets, and speed-based selection reaches its largest gain at a 20% budget. At this budget, speed-based selection and the configuration without the safety arbiter yield positive average improvements, whereas the remaining strategies stay close to zero. Inference deadlines further restrict which scenarios are eligible for Slow Planning: the eligible proportion rises sharply as the deadline approaches 22 s, but the budget still caps Slow Planning calls at 20%.

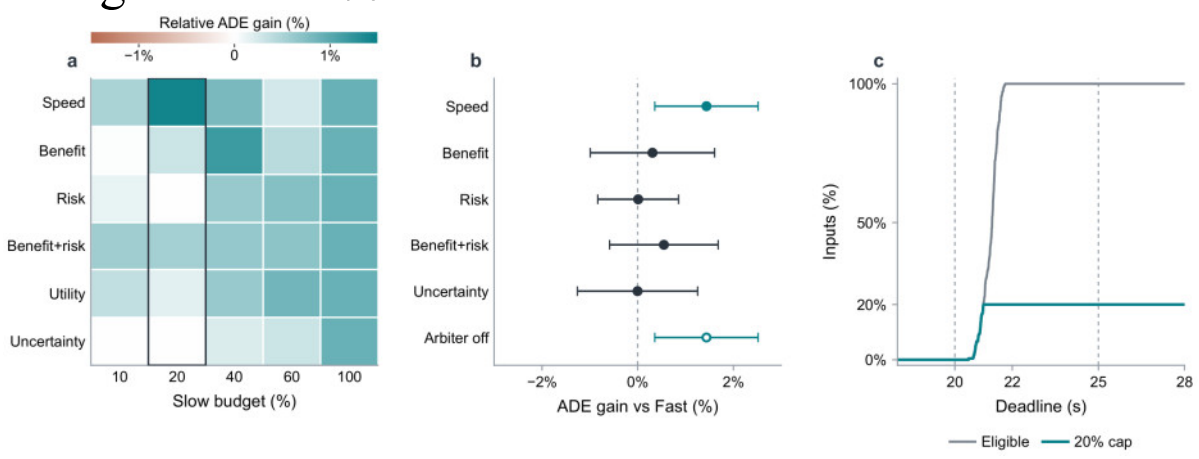

**Fig. 8.** Ablation analysis.

## F. Qualitative Analysis

The three representative scenarios in Fig. 9 show that explicit reasoning can improve, degrade, or leave unchanged the planning accuracy. In Case 01, a car-following scenario, Slow Planning produces a trajectory closer to the ground truth, reducing ADE from 1.496 m to 1.412 m and FDE from 4.348 m to 4.199 m. In Case 09, the reasoning recommends accelerating along the marked lane, but the resulting Slow Planning trajectory deviates further from the ground truth, increasing ADE from 1.157 m to 1.240 m. In Case 15, Fast and Slow Planning generate nearly identical trajectories, and both reach an ADE of 1.510 m. Such variations underscore the need to predict the reasoning benefit before invoking explicit reasoning.

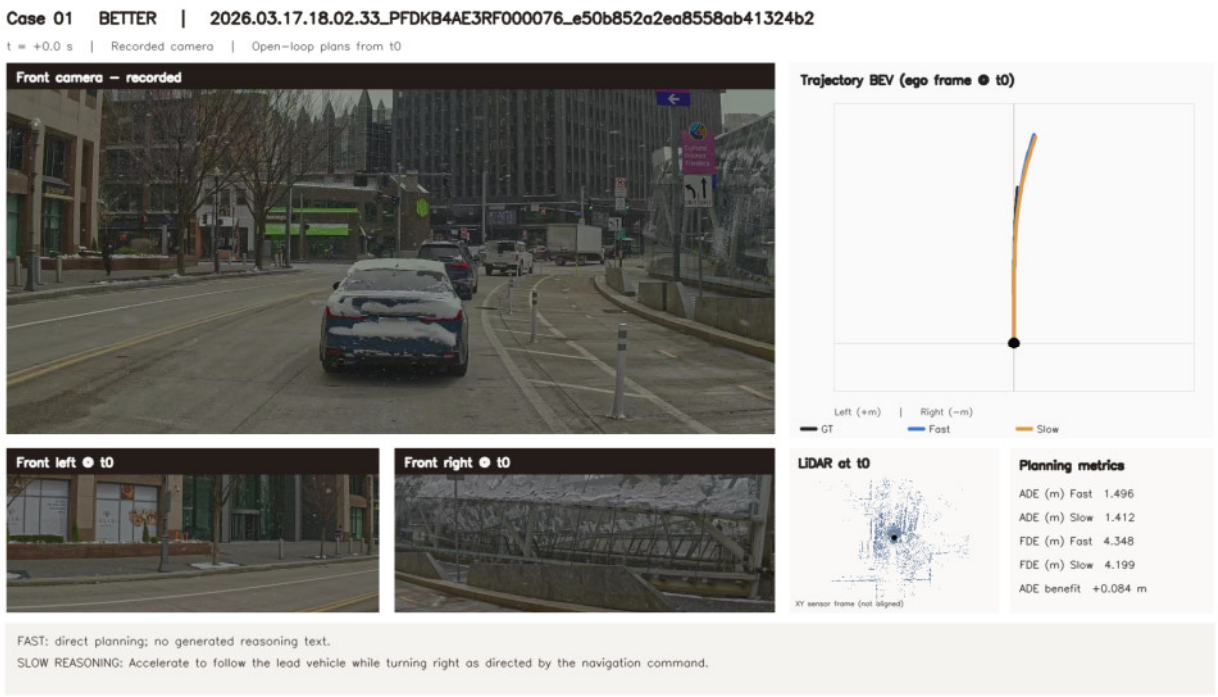

(*a*) Improvement

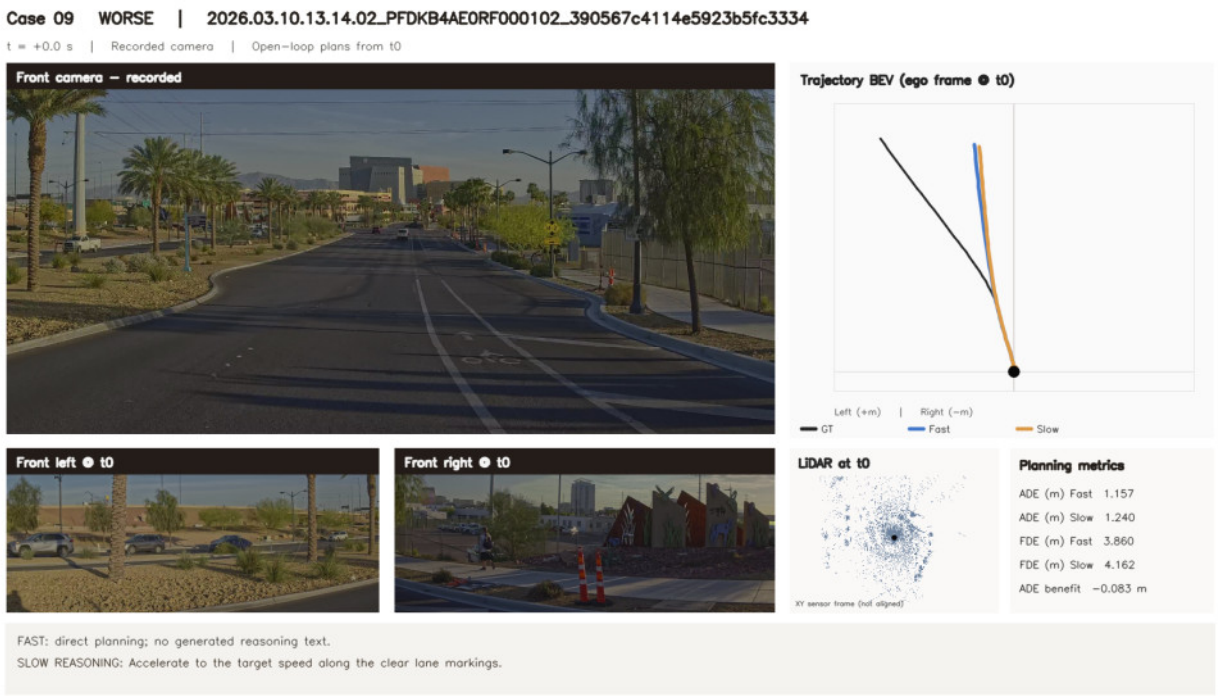

(*b*) Degradation

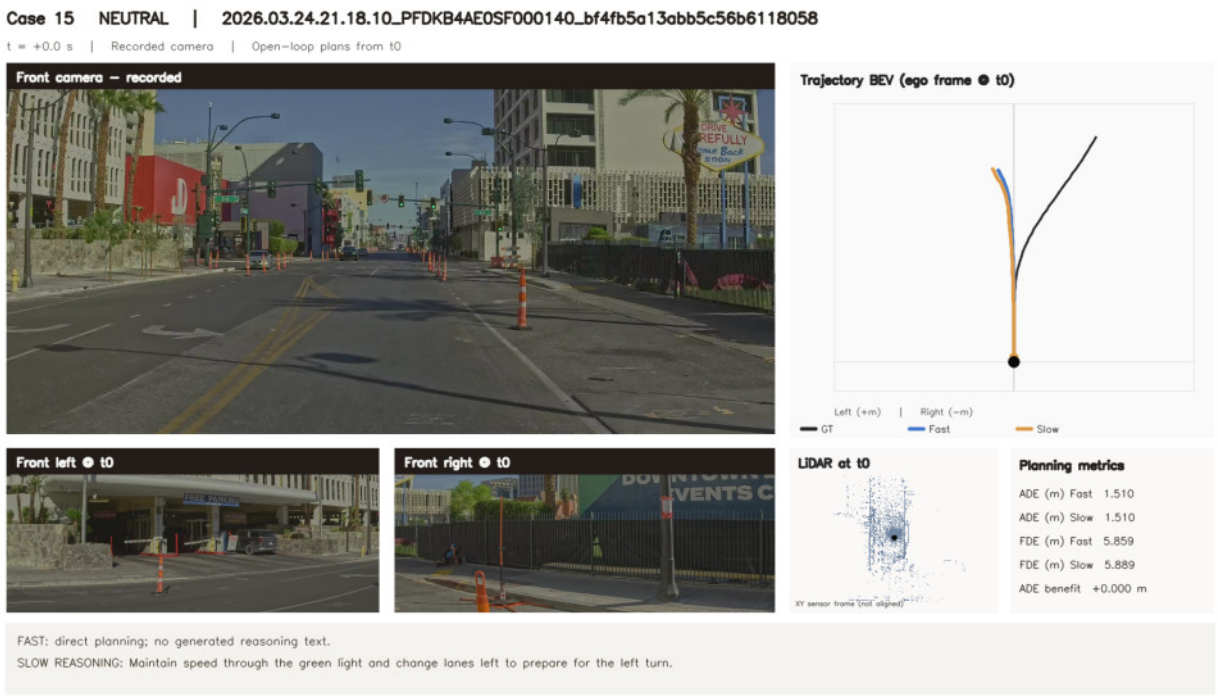

(*c*) No change

**Fig. 9.** Representative cases.

## V. Conclusion

This study investigated the effects of explicit reasoning on autonomous driving trajectory planning and proposed SAGE-Drive, a benefit-aware adaptive reasoning framework for autonomous driving planning. Analysis of 1,497 paired planning outcomes revealed that explicit reasoning reduced trajectory error in 53.5% of scenarios and increased it in 46.5%, resulting in a near-zero average benefit. To account for these variations, SAGE-Drive predicts the reasoning benefit from the Fast Planning trajectory and multimodal scene information and invokes Slow Planning only for the scenarios with the highest predicted benefits. Experimental results showed that SAGE-Drive achieved an ADE50 of 1.5191 m with a 20% reasoning budget, compared with 1.5252 m for Random Selection. At a 35% budget, SAGE-Drive achieved an ADE50 of 1.5148 m, close to the 1.5141 m of Slow-only, which invokes Slow Planning for every scenario. These findings show that benefit-aware reasoning selection retains most of the accuracy gain of explicit reasoning with a fraction of the Slow Planning calls. Future research will focus on real-time reasoning selection and closed-loop planning evaluation in dynamic traffic environments.

## Appendix

### *Appendix A: Reasoning benefit analysis*

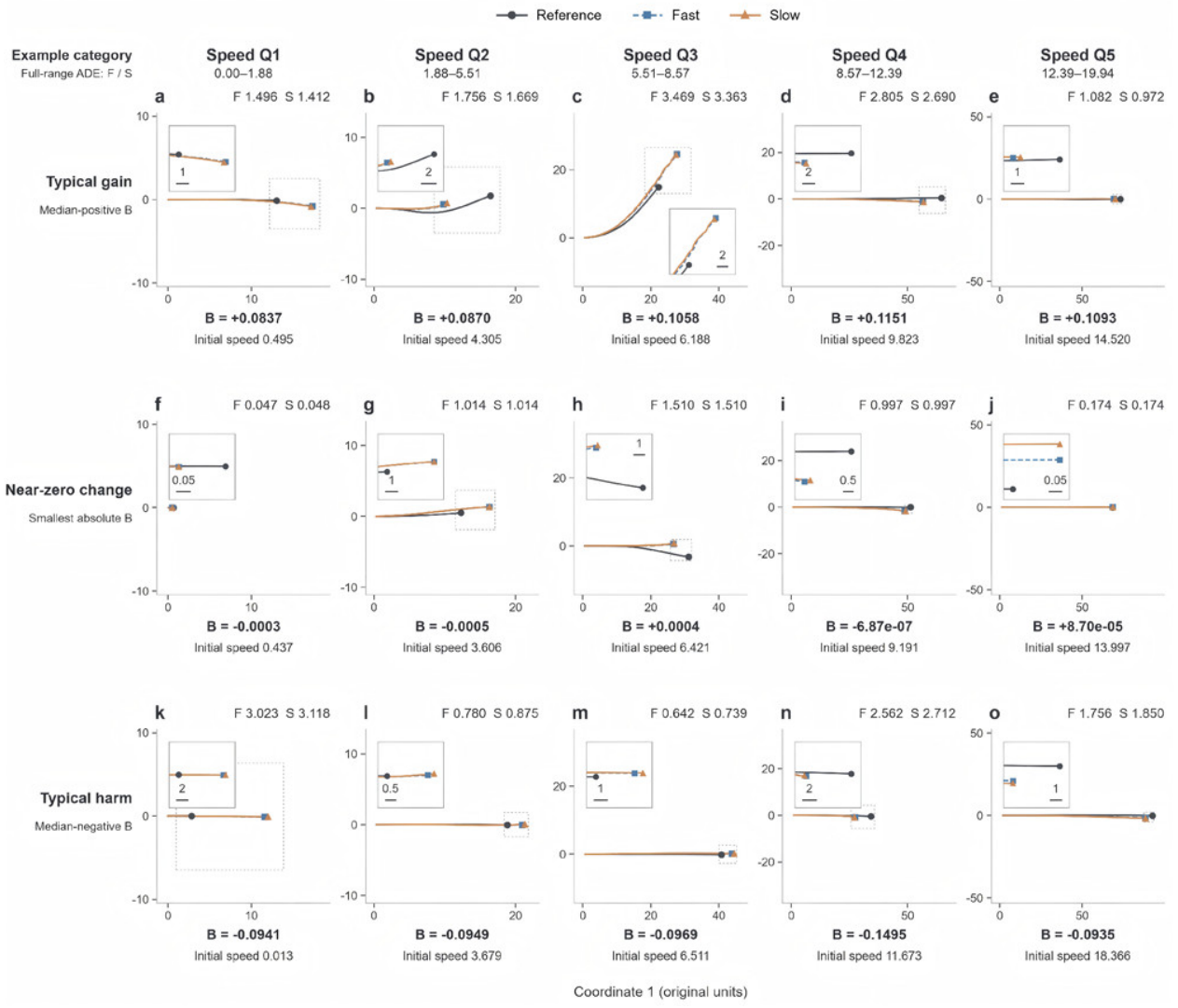

**Fig. A1.** Representative trajectory changes.

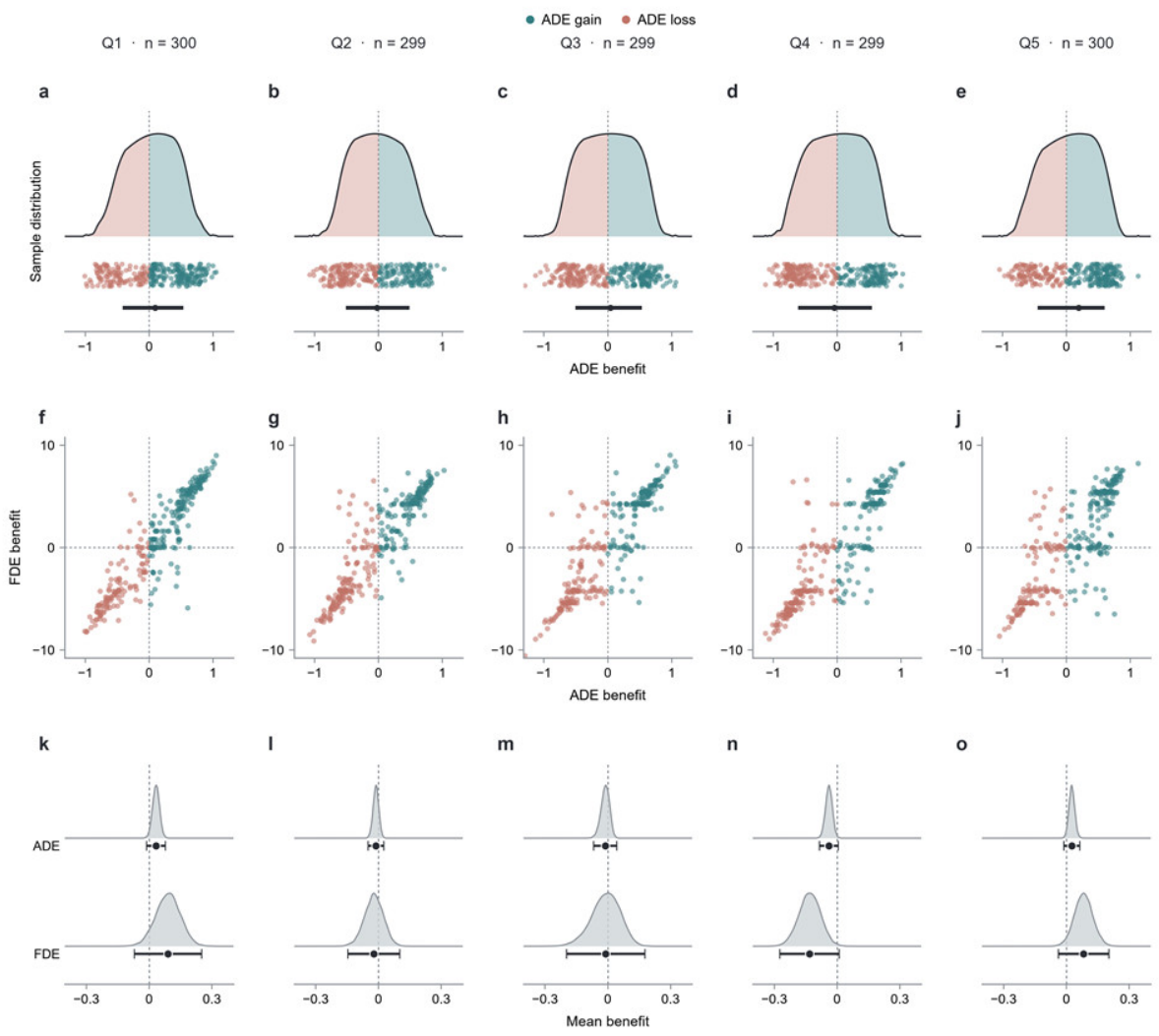

**Fig. A2.** Statistical analysis.

### *Appendix B: Retrieval-augmented reasoning analysis*

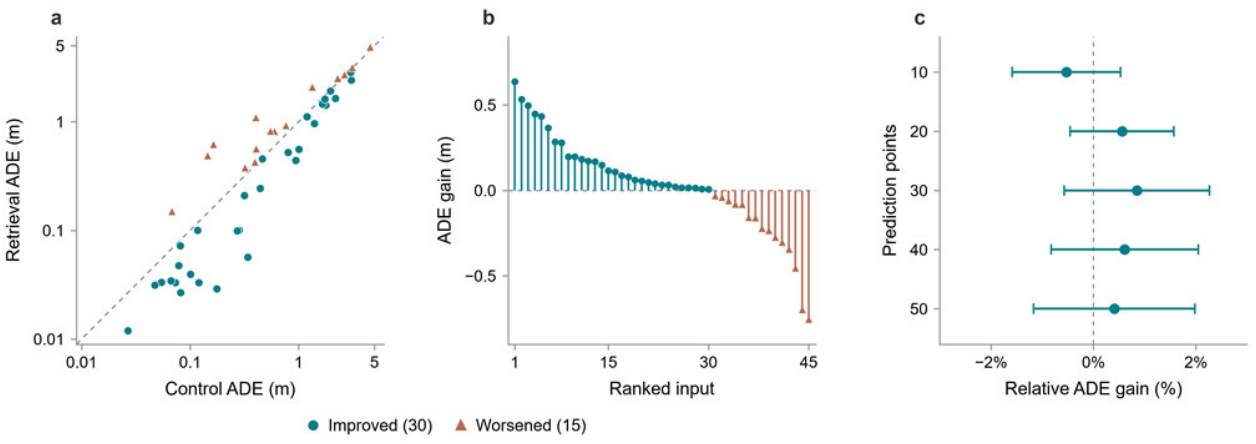

**Fig. B1.** Retrieval-augmented reasoning analysis.

## ACKNOWLEDGMENT

This study was supported by the National Key R&D Program of China (52672449).

## REFERENCES

[1] C. Sima et al., "DriveLM: Driving with graph visual question answering," in *Proc. Eur. Conf. Comput. Vis. (ECCV)*, 2024, pp. 256–274, doi: 10.1007/978-3-031-72943-0_15.

[2] M. Nie et al., "Reason2Drive: Towards interpretable and chain-based reasoning for autonomous driving," in *Proc. Eur. Conf. Comput. Vis. (ECCV)*, 2024, pp. 292–308, doi: 10.1007/978-3-031-73347-5_17.

[3] X. Tian et al., "DriveVLM: The convergence of autonomous driving and large vision-language models," in *Proc. Conf. Robot Learn. (CoRL)*, vol. 270, 2025, pp. 4698–4726.

[4] F. Kong et al., "VLR-Driver: Large vision-language-reasoning models for embodied autonomous driving," in *Proc. IEEE/CVF Int. Conf. Comput. Vis. (ICCV)*, 2025, pp. 26966–26975.

[5] G. Huang, T. She, Y. Lu, Y. Huang, Y. Ye, and Y. Shen, "FactorDrive: Adaptive multi-step reasoning driven by planning-critical factors for end-to-end autonomous driving," *arXiv preprint arXiv:2608.09591*, 2026.

[6] Y. Zou et al., "STEER: Route-aware adaptive reasoning for autonomous driving," *arXiv preprint arXiv:2611.00718*, 2026.

[7] T. Deruyttere, S. Vandenhende, D. Grujicic, L. Van Gool, and M.-F. Moens, "Talk2Car: Taking control of your self-driving car," in *Proc. EMNLP-IJCNLP*, 2019, pp. 2088–2098, doi: 10.18653/v1/D19-1215.

[8] T. Qian, J. Chen, L. Zhuo, Y. Jiao, and Y.-G. Jiang, "nuScenes-QA: A multi-modal visual question answering benchmark for autonomous driving scenario," in *Proc. AAAI Conf. Artif. Intell.*, vol. 38, no. 5, 2024, pp. 4542–4550, doi: 10.1609/aaai.v38i5.28253.

[9] A.-M. Marcu et al., "LingoQA: Visual question answering for autonomous driving," in *Proc. Eur. Conf. Comput. Vis. (ECCV)*, 2024, pp. 252–269, doi: 10.1007/978-3-031-72980-5_15.

[10] K. Chen et al., "Automated evaluation of large vision-language models on self-driving corner cases," in *Proc. IEEE/CVF Winter Conf. Appl. Comput. Vis. (WACV)*, 2025, pp. 7806–7815, doi: 10.1109/WACV61041.2025.00759.

[11] K. Tian, J. Mao, Y. Zhang, J. Jiang, Y. Zhou, and Z. Tu, "NuScenes-SpatialQA: A spatial understanding and reasoning benchmark for vision-language models in autonomous driving," in *Proc. IEEE/CVF Int. Conf. Comput. Vis. Workshops (ICCVW)*, 2025, pp. 4626–4635.

[12] S. Wang et al., "OmniDrive: A holistic vision-language dataset for autonomous driving with counterfactual reasoning," in *Proc. IEEE/CVF Conf. Comput. Vis. Pattern Recognit. (CVPR)*, 2025, pp. 22442–22452.

[13] E. Cui et al., "DriveMLM: Aligning multi-modal large language models with behavioral planning states for autonomous driving," *Vis. Intell.*, vol. 3, Art. no. 22, 2025, doi: 10.1007/s44267-025-00095-w.

[14] Y. Lu, J. Tu, Y. Ma, and X. Zhu, "ReAL-AD: Towards human-like reasoning in end-to-end autonomous driving," in *Proc. IEEE/CVF Int. Conf. Comput. Vis. (ICCV)*, 2025, pp. 27783–27793.

[15] J. Mao, Y. Qian, J. Ye, H. Zhao, and Y. Wang, "GPT-Driver: Learning to drive with GPT," in *NeurIPS 2023 Workshops: FMDM*, 2023.

[16] J.-J. Hwang et al., "EMMA: End-to-end multimodal model for autonomous driving," *Trans. Mach. Learn. Res.*, 2025.

[17] H. Fu et al., "ORION: A holistic end-to-end autonomous driving framework by vision-language instructed action generation," in *Proc. IEEE/CVF Int. Conf. Comput. Vis. (ICCV)*, 2025, pp. 24823–24834.

[18] Y. Li, M. Tian, D. Zhu, J. Zhu, Z. Lin, Z. Xiong, and X. Zhao, "Drive-R1: Bridging reasoning and planning in VLMs for autonomous driving with reinforcement learning," in *Proc. AAAI Conf. Artif. Intell.*, vol. 40, no. 8, 2026, pp. 6708–6716, doi: 10.1609/aaai.v40i8.37602.

[19] R. Zhang et al., "AdaDrive: Self-adaptive slow-fast system for language-grounded autonomous driving," in *Proc. IEEE/CVF Int. Conf. Comput. Vis. (ICCV)*, 2025, pp. 5112–5121.

[20] Z. Zhou et al., "AutoVLA: A vision-language-action model for end-to-end autonomous driving with adaptive reasoning and reinforcement fine-tuning," in *Adv. Neural Inf. Process. Syst. (NeurIPS)*, vol. 38, 2025, doi: 10.52202/085713-0942.

[21] Y. Luo et al., "AdaThinkDrive: Adaptive thinking via reinforcement learning for autonomous driving," in *Proc. IEEE Int. Conf. Robot. Autom. (ICRA)*, 2026, pp. 14489–14496, doi: 10.1109/ICRA57385.2026.11696257.

[22] B. Yang, B. Yao, T. Fu, and H. Du, "ThinkDrive: Adaptive dual-process reasoning for autonomous driving via uncertainty-triggered causal deliberation," *Mathematics*, vol. 14, no. 11, Art. no. 1806, 2026, doi: 10.3390/math14111806.

[23] X. Chen et al., "CAR-VLA: Complexity-aware and risk-adaptive reasoning for autonomous driving," *arXiv preprint arXiv:2609.34387*, 2026.

[24] M. Foutter, J. Dao, Y. Wang, D. Gammelli, and M. Pavone, "GovernorVLA: Adaptive embodied reasoning for long-tail autonomous driving," *presented at the ICRA Workshop on Semantics for Reliable Robot Autonomy*, 2026.

[25] S. Ang et al., "ASSCG: Just-right gating over chattering for fast-slow LLM planning in autonomous driving," *arXiv preprint arXiv:2606.25509*, 2026.

[26] X. Pei et al., "SlackDrive: Reclaiming runtime slack for adaptive driving inference," *arXiv preprint arXiv:2609.28064*, 2026.

[27] A. Bardes, J. Ponce, and Y. LeCun, "VICReg: Variance-invariance-covariance regularization for self-supervised learning," in *Proc. Int. Conf. Learn. Represent. (ICLR)*, 2022.

[28] Z. Huang et al., "nuReasoning: A reasoning-centric dataset and benchmark for long-tail autonomous driving," in *Advances in Neural Information Processing Systems (NeurIPS)*, 2026.

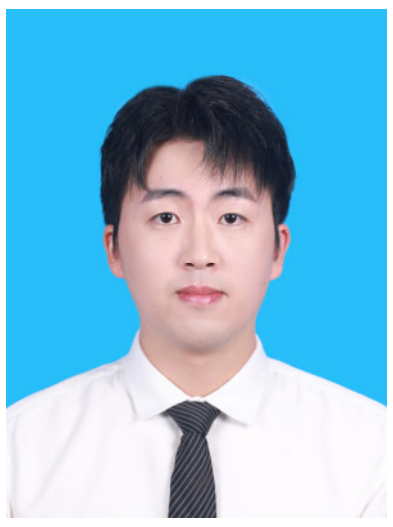

**Donglei Rong** received the Ph.D. degree in Transportation from Zhejiang University, Hangzhou, China, in 2026. He is currently a Postdoctoral Researcher with the School of Civil and Environmental Engineering, Georgia Institute of Technology, Atlanta, GA, USA. He is also pursuing a Ph.D. degree in Intelligent Transportation at The Hong Kong Polytechnic University, Hong Kong, through a dual-degree doctoral program. His research interests include transportation safety and driving behavior, multi-agent prediction and planning, end-to-end autonomous driving, and trustworthy artificial intelligence for intelligent transportation systems.

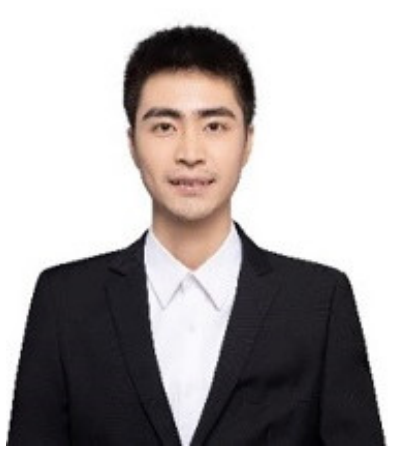

**Wenbin Yao** received the B.S. degree and Ph.D. degree from Zhejiang University. He is currently a lecturer with the School of Civil Engineering and Architecture, Zhejiang Sci-Tech University. His research interests include traffic demand management, and data-driven transportation system modeling and analysis.

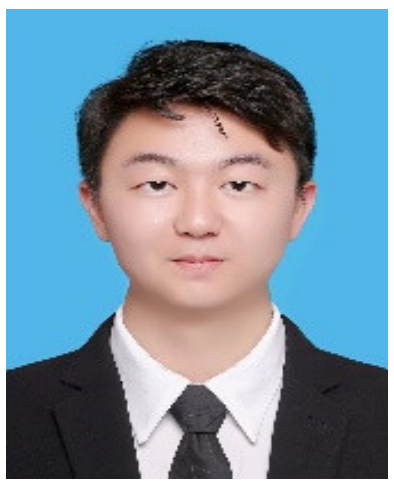

**Congcong Bai** recieved the B.S. degree in the Institute of Intelligent Transportation Systems of the College of Civil Engineering and Architecture, Zhejiang University, Hangzhou, Zhejiang, China, 2021. He is currently working toward the Ph.D degree with the Institute of Intelligent Transportation Systems of the College of Civil Engineering and Architecture, Zhejiang University, Hangzhou, Zhejiang, China. His research interest including micro driving behavior modeling, automatic driving behavior prediction and traffic big data mining.

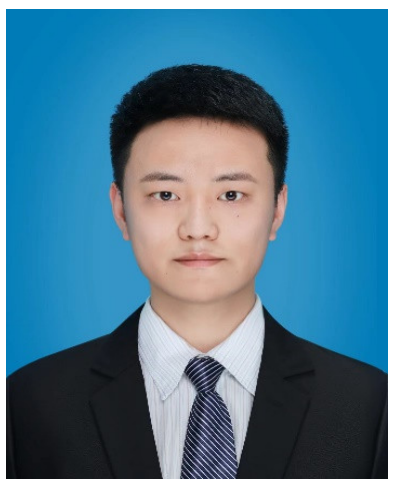

**Xi Gao** received the B.S. degree in the Transportation College of Jilin University, Changchun, Jilin, China, in 2023. He is currently working toward the Ph.D. degree with the Institute of Intelligent Transportation Systems of the College of Civil Engineering and Architecture, Zhejiang University, Hangzhou, Zhejiang, China. His research interest includes electric vehicle behavior and transportation big data.

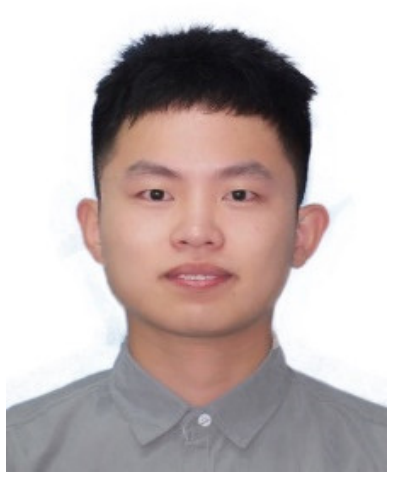

**Qiaoqiao Ren** received the bachelor's degree from Central South University and the master's degree from Southeast University in 2019 and 2022, respectively. He is currently pursuing the Ph.D. degree in the Department of Industrial and Systems Engineering, The Hong Kong Polytechnic University. His research interests include traffic safety analysis, autonomous vehicles, driver behavior, vehicle micro trajectory analysis, and applications of machine learning and deep learning in transportation engineering.

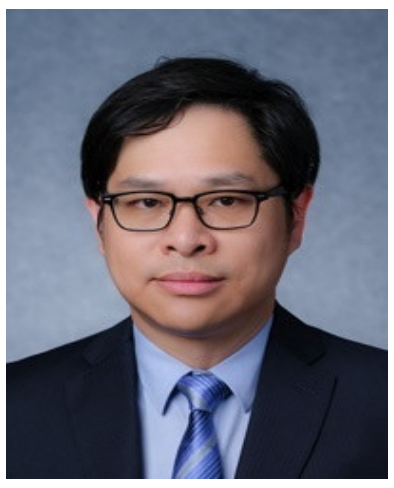

**Sheng Jin** received the Ph.D. degree in Transportation Engineering from Jilin University, Changchun, China, in 2010. Now, he is a Professor with the College of Civil Engineering and Architecture, Zhejiang University, Hangzhou, China. He has published more than 80 articles in related journals such as Transportation Research part A/B/C/D, Accident Analysis & Prevention, IEEE T ITS, and IEEE ITSM. His research is focused on traffic flow theory, intelligent transportation systems, traffic signal control, and traffic big data.

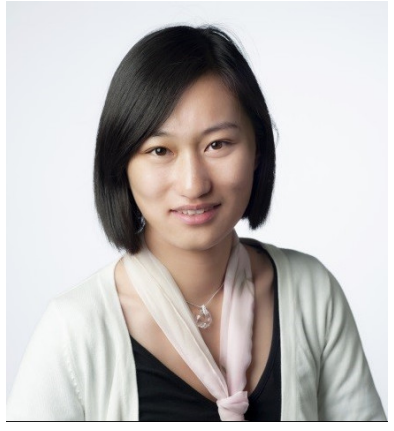

**Cheng Xu** is currently an Associate Professor in the Department of Traffic Management Engineering at Zhejiang Police College, Hangzhou, China. Her research interests include traffic flow theory and modeling, transportation planning, traffic safety, and intelligent transportation systems, with particular emphasis on mixed traffic flow, non-motorized transportation, and traffic risk analysis.